\pdfoutput=1

\documentclass[11pt]{article}

\usepackage[]{acl}

\usepackage{times}
\usepackage{latexsym}

\usepackage[T1]{fontenc}

\usepackage[utf8]{inputenc}

\usepackage{hyperref}

\usepackage{adjustbox}

\usepackage{pifont}
\newcommand{\xmark}{\ding{55}}%
\usepackage{amssymb} 

\usepackage{booktabs}
\usepackage{array}

\usepackage{microtype}
\usepackage{todonotes}

\usepackage{enumerate}

\title{Contextual Sentence Classification:\\ Detecting Sustainability Initiatives in Company Reports}

  \author{Dan Hirlea$^{1}$  \and  Christopher Bryant$^{3}$ \and Maurizio Zollo$^{2}$ \and Marek Rei$^{1,2,3}$\\
$^1$Department of Computing, Imperial College London, UK \\
$^2$Leonardo Centre, Imperial College London, UK  \\
$^3$ALTA Institute, Computer Laboratory, University of Cambridge, UK \\
  {\tt \footnotesize mh120@ic.ac.uk, cjb255@cl.cam.ac.uk, m.zollo@imperial.ac.uk,  marek.rei@imperial.ac.uk}}

\begin{document}
\maketitle
\begin{abstract}
We introduce the novel task of detecting sustainability initiatives in company reports. Given a full report, the aim is to automatically identify mentions of practical activities that a company has performed in order to tackle specific societal issues.
New methods for identifying continuous sentence spans need to be developed for capturing the multi-sentence structure of individual sustainability initiatives.
We release a new dataset of company reports in which the text has been manually annotated with sustainability initiatives. We also evaluate different models for initiative detection, introducing a novel aggregation and evaluation methodology. Our proposed architecture uses sequences of consecutive sentences to account for contextual information when making classification decisions at the individual sentence level.

\end{abstract}


\section{Introduction}

Sustainability has become an important theme around the world as global resource consumption continues to increase, driven by population and economic growth. In an effort to meet these challenges, many companies release annual reports which describe everything they have accomplished during the year, including any steps they have taken towards sustainable development.
Automated models for the detection of such sustainability initiatives allow us to construct company-level and sector-level sustainability profiles at scale, across potentially hundreds of thousands of reports. In the future, this opens up the possibility of correlating these metrics with other information, such as financial data, to predict which initiatives each company is likely to take, or to predict which types of initiatives would be most optimal in terms of revenue or public support.

The automated detection of these mentions is a challenging NLP task, as it often requires full semantic understanding of the text combined with some background knowledge. Not all actions described in these reports should be classified as valid initiatives; for example, some may refer to plans that will be implemented in the future, while others can be so vague as to not describe a specific achievement. A sustainability initiative is thus defined as \textit{a practical activity or set of related activities that a firm performs in order to tackle a societal issue}. Some examples of valid (\checkmark) and invalid (\xmark) sustainability initiatives are shown below: 

\begin{enumerate}[(a)]
\setlength\itemsep{-0.1cm}
\small
     \item  \textit{We will build and operate a new, highly efficient steam generator for the chemical company Dow Benelux in the coming year.} (\xmark) 
     \item \textit{This year, we further increased employee satisfaction by 2\%.} (\xmark) 
     \item \textit{This year, we further increased employee satisfaction by 2\%, launching a new career management program.} (\checkmark) 
     \item \textit{Through our projects in industry and commerce, our B2B-customers have made a total of 2,215 GWh of energy savings in 2017.} (\xmark) 
     \item \textit{Through our projects in industry and commerce, our B2B-customers have made a total of 2,215 GWh of energy savings in 2017. We owe these savings mainly to the increase in fuel efficiency in the B2B business and the expansion of digital solutions.} (\checkmark) 
\end{enumerate}

In these examples, (a) is invalid because it describes a future action that has not yet been implemented, while (b) is invalid because it describes an achievement without a specified action. In contrast, (c) is valid because it includes the sustainability action that was taken that led to the achievement. Similarly, (d) does not specify a particular action, while (e) states the actions that were taken. (e) is also an example of a multi-sentence initiative, which highlights the fact that it is insufficient to classify sentences in isolation when detecting sustainability initiatives. Further examples alongside a summary of the annotation guidelines are provided in \autoref{sec:guidelines}. 

In this work, we release a new dataset, containing 192,978 sentences from 45 company reports, that have been manually annotated with sustainability initiatives. We hope that this dataset, which we make publicly available, will serve as a benchmark for future work on sustainability initiative detection. 

We also investigate the performance of different neural models on this dataset and propose a novel architecture and evaluation methodology specifically tailored to the initiative detection task. The best model is based on RoBERTa \citep{liu2019roberta} and jointly encodes a sequence of multiple sentences, on each side of the target sentence, in order to provide further context for the decision process. Fine-grained labels are then predicted for each of the input sentences using a CRF \citep{lafferty2001}, which increases the coherence of the consecutive predictions while also providing a multi-task objective to the model. Fine-grained sentence-level predictions are then used to construct continuous spans describing each sustainability initiative.

The dataset, along with usage instructions and our code base are available here.\footnote{\url{https://github.com/dhirlea/contextual_sentence_classification}}


\section{Background}

In this paper, we employ pre-trained Transformer models \citep{vaswani2017attention}, specifically BERT \citep{devlin2019bert} and RoBERTa \citep{liu2019roberta} as implemented in the HuggingFace Transformers library \citep{wolf-etal-2020-transformers}, to perform sequence classification at both the single-sentence and multi-sentence level. The decision to use these models is based on their ability to perform well on text classification tasks, e.g. GLEU \citep{wang2018glue}, which illustrates their strong natural language understanding capability.

The task of initiative detection requires aggregating information across multiple sentences. \citet{beltagy2020longformer} introduced the Longformer model which accepts as input sequences of documents separated by a special end of document token. \citet{dasigi2021dataset} on the other hand join together several sentences using a special separation token in order to perform text classification. Similar to both of their work, we design a transformer model that accepts as input sequences of multiple sentences separated by a special separation token. However, we also incorporate multiple classification heads, corresponding to each of the input sentences, thereby introducing additional contextual information about the surrounding output labels.

\citet{luoma2020exploring} showed that multi-sentence contextualisation is able to improve model performance. They explore using cross-sentence context for Named Entity Recognition (NER) firstly by establishing a baseline to predict tags for all tokens within individual sentences. They then augment the process to construct input examples consisting of several input sentences and move the position of the sentence containing the tokens of interest across the context window to explore whether changing the level of context improves classification performance. Lastly, majority voting is used to predict tags for all tokens within a given sentence seen in different contexts at inference time. The authors conclude that including multi-sentence context helps improve results for BERT-based NER tasks in English, Dutch and Finnish text. While our task requires classification at the sentence level, as opposed to labeling individual tokens, this finding motivates the use of multi-sentence context to aid the classification decision of our system.

In order to improve the sentence-level predictions of the model, we decode the labels of the contextualised sentences using a CRF layer \cite{lafferty2001}, similar to the model for propaganda detection in news articles proposed by \citet{chauhan-diddee-2020-psuedoprop}. In contrast to their system, we first predict sustainability labels for individual sentences and then use a post-processing module to aggregate sentence labels across consecutive sentences into full initiative spans, whereas the model by \citet{chauhan-diddee-2020-psuedoprop} first predicts sentences of interest, then zooms into these sentences to identify the token-level span of individual propaganda elements. 

When aggregating predictions across several sentences, we also make use of the IOBES schema \citep{Krishnan2005NamedER,lester2020iobes} inspired by the CoNLL-2003 shared task on Named Entity Recognition \citep{sang2003introduction}. Specifically, in NER, each token in a dataset can be tagged as I (Inside), O (Outside), B (Beginning), E (End), or S (Singleton) and these labels are aggregated into named entities spanning one (e.g. S) or multiple tokens (e.g. BIIE). We apply the same methodology to sustainability initiatives, except do so at the sentence level rather than the token level. This is one of the novel contributions of our work that ensures we predict coherent sustainability initiatives across multiple consecutive sentences.

Finally, event coreference resolution  \citep{lu-ng-2018,zeng-etal-2020-event} is a related task that also deals with the detection of entity mentions in documents. However, the task of detecting sustainability initiatives requires identifying \textit{consecutive} sentences that form \textit{unique} initiatives in a \textit{single} document, rather than \textit{disjoint} sentences that refer to the \textit{same} events in \textit{multiple} documents. Consequently, many of the sentence similarity techniques used in event coreference resolution research do not apply to our task, given that initiatives usually consist of semantically dissimilar sentences that are unrelated to other initiatives.


\section{Dataset}
\label{sec:Dataset}

We introduce a novel dataset, containing 192,978 sentences from 45 company reports,  which were annotated with distinct sustainability initiatives by annotators from the Leonardo Centre for Sustainable Business. Specifically, annotators read through PDF reports published by different companies and extracted text sections that corresponded to sustainability initiatives. Each PDF was then converted to plain text and the extracted text sections were mapped back into the full report.  The conversion was performed using the \textit{pdftotxt} tool of the Poppler library\footnote{\url{https://poppler.freedesktop.org/releases.html}} and individual sentences were extracted using spaCy\footnote{\url{https://spacy.io/}}. The dataset has been split into training (87,469 sentences from 25 reports), development (55,630 sentences from 10 reports) and test (55,630 from 10 reports) sets. Statistics about the dataset splits are shown in \autoref{table:data_overview}.  

Many of the sustainability initiatives in our dataset are associated with sustainable development goals (SDGs) as defined by the United Nations in 2015.\footnote{\url{https://sdgs.un.org/goals}} SDGs are general targets that the United Nations believe all economies should aim for in order to support the Circular Economy. These SDGs act as additional topic-related labels that provide possibilities for interesting multi-task optimization experiments in future work.

\begin{table}[t]\centering
\begin{adjustbox}{width=0.48\textwidth}
\begin{tabular}{@{}lccc@{}}\toprule
\multicolumn{1}{c}{}& \multicolumn{1}{c}{Train} & \multicolumn{1}{c}{Develop} & \multicolumn{1}{c}{Test}\\
\midrule
\# reports &  25 & 10 & 10 \\
\# sentences &  87,469 & 55,630 & 49,879\\
\# initiatives & 2,040 & 985 & 1,024\\
\% sentences with initiatives & 4.19\% & 3.19\% & 3.42\%\\
\% initiatives with SDGs & 80.00\% &52.08\% & 72.95\%\\
\bottomrule
\end{tabular}
\end{adjustbox}
\caption{General statistics about the dataset.}
\label{table:data_overview}
\end{table}


\subsection{Dataset Reliability}

\begin{table}\centering
\begin{adjustbox}{width=0.48\textwidth}
\begin{tabular}{@{}lrrccc@{}}\toprule
Report& A1 & A2 & \# Matches & Min \% & Max \%\\
\midrule
Cosmote 2008 & 144 & 133 & 102 & 58.29\% & 76.69\% \\
Cosmote 2009 & 138 & 87 & 82 & 57.34\% & 94.25\% \\
Portel 2008 & 177 & 176 & 139 & 64.95\% & 78.98\% \\
TeliaSonera 2008 & 102 & 97 & 65 & 48.51\% & 67.01\% \\
TeliaSonera 2009 & 117 & 129 & 58 & 30.85\% & 49.57\% \\
\midrule
Average & & & & 51.99\% & 73.30\%\\
\bottomrule
\end{tabular}
\end{adjustbox}
\caption{Inter-annotator agreement results reproduced for the dataset. A1 and A2 indicate the number of initiatives returned by each annotator. \textit{\# Matches} shows the number of initiatives that were returned by both annotators. Min/Max percentages are calculated using \autoref{eq:min_match} and \autoref{eq:max_match}. 
}
\label{table:annotator}
\end{table}

Although each report in the dataset is  annotated by a single annotator, the Leonardo Centre independently carried out a small-scale study on inter-annotator agreement to verify the integrity of the annotations \cite{leonardo2014}. Specifically, 5 company reports were selected and double-annotated in order to compare the initiatives that were extracted by each annotator.

We reproduce the results of this study in \autoref{table:annotator}, along with the number of initiatives detected by each annotator and the total number of initiatives agreed upon by both annotators. Initiative matching for this study was performed manually using partial textual overlap between the initiative span identified by the first annotator relative to the span identified by the second annotator. 

Since initiative detection is a structured prediction task (each initiative can contain a variable number of sentences), Cohen's kappa cannot be directly calculated and we instead measure the minimum and maximum match percentage between the annotators.
The minimum estimated number of initiatives per report is the minimum number of paragraphs identified by either annotator. The maximum estimated number of initiatives is the union of the sets of paragraphs identified by both annotators. Using these estimates for the number of ground-truth initiatives per report, we define the match percentages between human labelers as per \autoref{eq:min_match} and \autoref{eq:max_match} in order to determine a range of agreement.

\begin{equation}\label{eq:min_match}
Match_{Min}(\%) = \frac{n_m}{n_1 + n_2 - n_m}
\end{equation}
\begin{equation}\label{eq:max_match}
Match_{Max}(\%) = \frac{n_m}{min(n_1, n_2)}
\end{equation}

\noindent $n_1$ and $n_2$ refer to initiatives found by annotators 1 and 2 respectively, and $n_m$ refers to the number of matching initiatives between the two annotators.

On average we can see that the agreement ranges between 52\% and 73\% which suggests that while there is meaningful agreement, there is also subjectivity in the task. Manual inspection of the dataset suggests that the differences in labeling across annotators can occur based on how much context a particular annotator decides to include. There are also false negative examples where single sentence initiatives were missed during labeling due to annotator fatigue when working through very long reports. However, the $Match_{Max}$ score shows that 73\% of the initiatives extracted by the more conservative annotator were also found by the second annotator, indicating a majority of high-confidence initiatives in the dataset.



\subsection{Data Exploration}

\begin{figure}[t]
\includegraphics[width=0.49\textwidth]{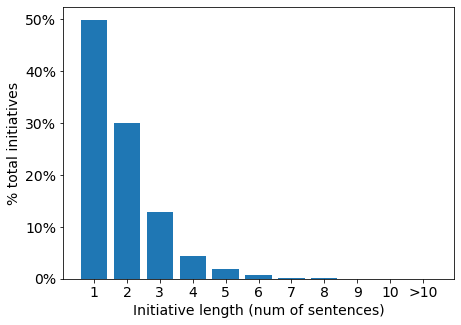}
\caption{Distribution of initiative length across training set}
\label{fig:init_length}
\end{figure}

As shown in \autoref{table:data_overview}, fewer than 4\% of sentences in our dataset are part of a sustainability initiative. This is because company sustainability reports are often very long texts that additionally contain sections on general company goals, stakeholder interactions, financial considerations, etc. which are peripheral to the task. This makes the task particularly challenging and motivates the use of neural models that are specialised to detect such a minority class in a highly imbalanced data distribution. We also analysed the distribution of initiatives in terms of sentence length. \autoref{fig:init_length} shows that just under 50\% of initiatives consist of a single sentence, and the vast majority of initiatives span no more than 4 sentences. This finding motivates the need to take context into account when detecting initiatives. 

A particular challenge of this dataset is the need to be able to differentiate between i) several adjacent initiatives which may be semantically similar and ii) a single multi-sentence initiative. A reliable system must therefore be able to correctly predict whether a sentence belongs to a multi-sentence initiative or whether it is a singleton initiative that is adjacent to another initiative. 


\section{Methodology}

The main aim of our work is to identify the precise spans of sentences that constitute distinct sustainability initiatives in a large number of company reports. We investigate the performance of standard pre-trained language models on this task, while also proposing modifications that improve their performance in this setting. The following sections describe the components of our system.


\subsection{Preprocessing}
\label{subsec:Preprocessing}

As previously mentioned in Section \ref{sec:Dataset}, the sentences in our dataset are extracted from PDFs using \textit{pdftotxt} and tokenized using \textit{spaCy}. This process is not perfect however, and some `sentences' consist of short strings extracted from cells in PDF tables, while others consist of long paragraphs of continuous text. Since these very short and very long sentences rarely form part of an initiative, we instantly classify all sentences shorter than 5 tokens or longer than 100 tokens as not part of an initiative in order to speed up processing time. These thresholds were tuned based on the development set. 


\subsection{Model Parameters}

All our models are trained using pre-trained BERT \citep{devlin2019bert} or RoBERTa \citep{liu2019roberta}. Specifically, we use the $\mathrm{BERT}_\mathrm{BASE}$ and $\mathrm{RoBERTa}_\mathrm{BASE}$ models implemented in the HuggingFace Transformers library \citep{wolf-etal-2020-transformers}. Models based on BERT make use of WordPiece tokenization \cite{wu2016googles} while models based on RoBERTa make use of Byte-Pair Encoding (BPE) tokenization \cite{sennrich2016neural}. We attach a fully connected layer using a dropout of 0.1 after the final encoding block of the Transformer. This is in line with standard fine-tuning methodology to help regularize the model and we train it using binary cross-entropy loss or log-likelihood loss depending on the use case.

Each model is trained for a maximum of 10 epochs and stops early if the F1 score on the positive \textit{is-initiative} class deteriorates by more than 3\% or there is no improvement after 5 epochs on the development set. We use a batch size of 16, a learning rate of 1e-5 with a warm-up over 6\% of the total training steps followed by linear decay. We made use of an Adam optimizer \citep{kingma2017adam} when fine-tuning the model. These hyperparameters are in line with those used in the original papers for BERT and RoBERTa and were chosen based on the performance of the models on the development set. We use these hyperparameters for all experiments throughout the paper.


\subsection{Binary vs. IOBES Labels}

In addition to comparing performance using different pre-trained models, we also compare performance using different sets of sentence labels. In the simplest case, we assign a binary label to each sentence and train a model to predict whether a sentence is part of an initiative or not. Adjacent sentences that are predicted to be part of an initiative are then aggregated and treated as a single multi-sentence initiative. The obvious limitation of this approach is that binary label aggregation is unable to differentiate between separate adjacent initiatives and multi-sentence initiatives. 

With this in mind, we also trained a multi-class classification system to assign IOBES labels to each sentence in our dataset. This approach was inspired by work on Named Entity Recognition \citep{sang2003introduction,Krishnan2005NamedER} which uses a similar scheme to assign part-of-entity tags at the token level. The main advantage of IOBES labels over binary labels is that they explicitly define the boundaries between initiatives and encode a finer-grained signal. For example, if an initiative consists of 4 consecutive sentences, the B(eginning) tag is assigned to the first sentence, the I(nside) tag is assigned to the second and third sentences and the E(nd) tag is assigned to the last sentence. The S(ingleton) tag meanwhile identifies initiatives that only consist of a single sentence while the O(utside) tag refers to all other sentences. Examples of these tags are shown in \autoref{table:sentence_labels}.

\begin{table}\centering
\begin{adjustbox}{width=0.48\textwidth}
\begin{tabular}{@{}cccc@{}}\toprule
\multicolumn{1}{c}{}& \multicolumn{1}{c}{Initiative ID} & \multicolumn{1}{c}{Binary Label} & \multicolumn{1}{c}{IOBES Label}\\
\midrule
Sentence 1 & - & 0 & O\\
Sentence 2 & NRGUS201501 & 1 & S\\
Sentence 3 & NRGUS201502 & 1 & B\\
Sentence 4 & NRGUS201502 & 1 & I\\
Sentence 5 & NRGUS201502 & 1 & I\\
Sentence 6 & NRGUS201502 & 1 & E\\
Sentence 7 & - & 0 & O\\
\bottomrule
\end{tabular}
\end{adjustbox}
\caption{The difference between binary and IOBES sentence labelling. Binary labels aggregate all sentences into a single initiative while IOBES labels differentiate between a singleton initiative and a BIIE initiative.}
\label{table:sentence_labels}
\end{table}

One complication of using IOBES labels over binary labels, however, is that there is no guarantee that the model will predict a coherent sequence of labels; for example, it might predict an End label without a Beginning. Knowing that most initiatives consist of a maximum of 5 sentences (\autoref{fig:init_length}), we aggregate sentence level predictions into the following initiative structures: Singletons (S), Beginning-End (BE), Beginning-Inside-End (BIE), Beginning-Inside-Inside-End (BIIE) and Beginning-Inside-Inside-Inside-End (BIIIE). If the model does not predict one of these structures, we treat all the relevant sentences as Singletons.


\subsection{Context}
\label{sec:context}

Given that sustainability initiatives typically consist of two parts: the initiative description and the initiative action, we hypothesize that adding context information around an input sentence will improve the model's predictive ability by allowing it to focus the attention mechanism on several potentially useful descriptive sentences. To give an example, the following extract from a report\footnote{\url{https://www.eon.com/content/dam/eon/eon-com/Documents/en/sustainability-report/EON_Sustainability_Report_2017.pdf}} shows a Singleton initiative sentence (bold) surrounded by one sentence of context either side. 

\begin{quote} 
    \small 
    \emph{
    "As part of our three-year plan for occupational health and safety, we initiated further new measures in these areas in 2017 and continued to implement existing ones. \textbf{These included, for example, special trainings for senior managers to enable them to better assess safety risks for their employees}. In connection with our three-year plan, we set up working groups at the beginning of 2016, which met four times in the reporting period."}
\end{quote}

\begin{figure*}[t!]
\includegraphics[width=\textwidth]{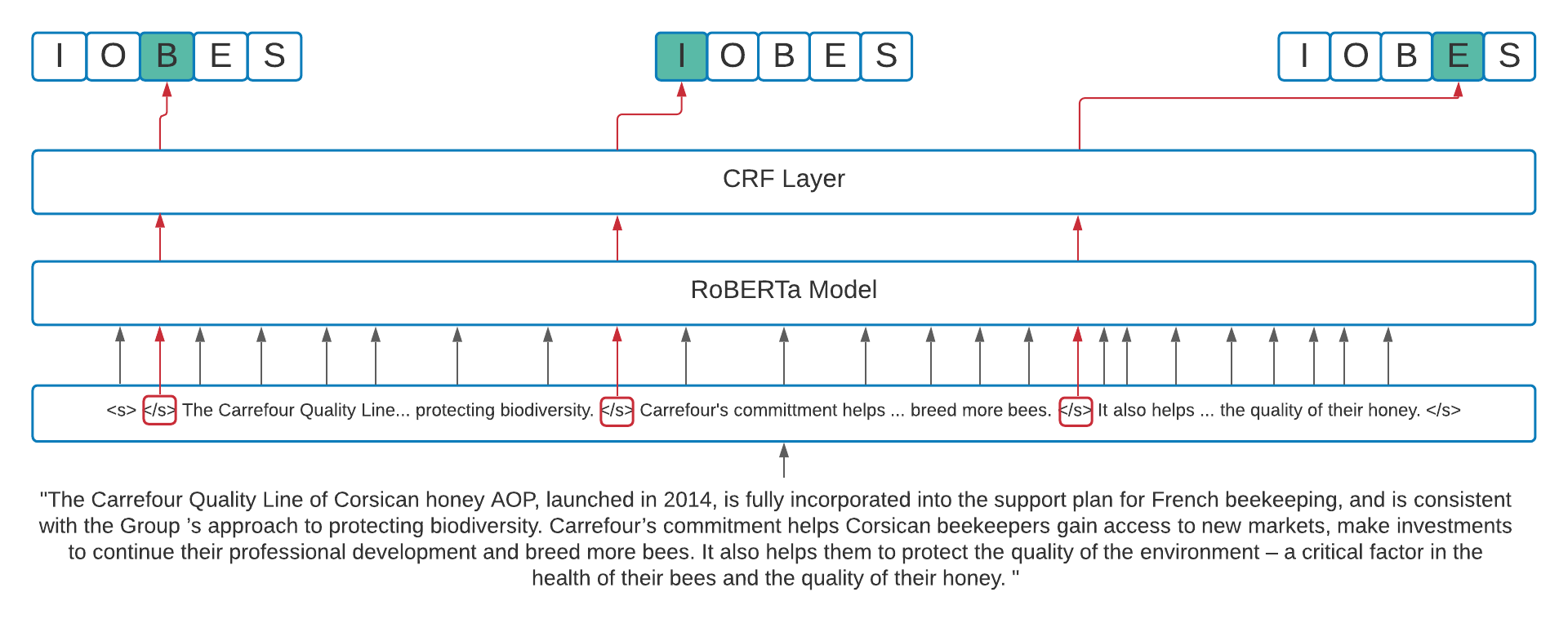}
\caption{Contextual sentence classification architecture for initiative detection. A sequence of 3 sentences is input to the model and a separate output label is predicted for each sentence, based on the representations of the separator tokens.}
\label{fig:system_architecture}
\end{figure*}

These contextual sentences are not part of the initiative, but still provide a strong indication that the middle sentence is an initiative. Specifically, the first sentence refers to vague ``new measures'' that were ``initiated'', which might suggest the following sentence will give a specific example, while the third sentence refers to the same ``three-year plan'' in the first sentence, which signals the entire discourse is on the same topic. A contextual model might thus be able to leverage this information to make a more informed decision about whether a sentence is part of an initiative or not. 

In order to add contextual information to our models, we concatenate three input sentences together (i.e. the target sentence with one sentence of context either side) with each sentence separated by a special separator token ([SEP] in BERT and </s> in RoBERTa) and a final separator token at the end denoting the end of the sequence. We then add a fully connected classification layer on top of each separator token at the start of a sentence. This allows the model to predict a label not only for the target sentence in the middle, but also for each of the context sentences. This combination  acts as a multi-task objective -- the main goal of the model is to predict the correct label for the target sentence, while the secondary objective is to also predict the labels for the supporting context sentences. Connecting the classifier heads to separator tokens allows the model to construct more informative representations of each individual sentence in the lower layers, instead of compressing everything into a single CLS representation.

During training with context, we consider the predictions of all the input sentences when computing the loss at the output of the network. However, at inference time we only record the prediction from the separator token corresponding to the target sentence in the middle of each input sequence -- this ensures that each prediction is made based on all the available context information. 
To further improve the model's predictive ability, we apply a CRF layer \citep{lafferty2001} on the output logits of the model, instead of directly using a softmax function when predicting IOBES tags. The CRF layer learns the transition probabilities of the labels and is therefore able to improve the coherence of neighbouring labels when predicting multi-sentence initiatives. An overview of the full system incorporating IOBES labels and context is shown in \autoref{fig:system_architecture}.


\subsection{Evaluation}

\begin{table*}[t!]\centering
\begin{tabular}{@{}lcccccc@{}}\toprule
& \multicolumn{3}{c}{Min Match} & \multicolumn{3}{c}{Exact Match} \\ 
\cmidrule(r){2-4}  \cmidrule(r){5-7}
\multicolumn{1}{c}{}& \multicolumn{1}{c}{Precision} & \multicolumn{1}{c}{Recall} & \multicolumn{1}{c}{F1} & \multicolumn{1}{c}{Precision} & \multicolumn{1}{c}{Recall} & \multicolumn{1}{c}{F1}\\ \midrule
BERT Binary Without Context & \textbf{39.45} & 44.73 & \textbf{41.92} & 14.64 & 16.60 & 15.56\\
BERT IOBES Without Context & 30.67 & 46.19 & 36.87 & 16.28 & 24.51 & 19.56\\
BERT IOBES With Context   & 32.41 & 46.97 & 38.36 & 16.91 & 24.51 & 20.02 \\
RoBERTa Binary Without Context   & 34.39 & 46.78 & 39.64 & 11.41 & 15.53 & 13.16 \\
RoBERTa IOBES Without Context   & 28.58 & 52.25 & 36.95 & 14.48 & 26.46 & 18.72 \\
RoBERTa IOBES With Context & 34.23 & \textbf{53.52}  & 41.75  & \textbf{19.30} & \textbf{30.18} & \textbf{23.54}\\
\bottomrule
\end{tabular}
\caption{Final system initiative level F1 scores on the test set. Several intermediary architecture options are shown to facilitate comparison and to support design choices.}
\label{table:final_results}
\end{table*}

We evaluate system performance using precision, recall and F1-score. Given that the task is to identify a textual span which can potentially consist of one or more sentences, there are multiple ways to define a match between a prediction and a target initiative. 
We therefore define two different metrics.
The \textbf{Min Match} metric is somewhat lenient and rewards the system if at least one sentence in the predicted initiative overlaps with the target span. 
The \textbf{Exact Match} metric is more strict and rewards the system only if both the start and end points of the predicted initiative exactly match the target span. 
The importance of each metric depends on the application -- the Min Match metric reflects how accurately the system can detect the approximate location of an initiative, which can later be reviewed by a human user, while the Exact Match metric reflects how accurately the system performs the task in a completely automated way.


\section{Results and Analysis}

We present the results of all our experiments in \autoref{table:final_results}. Focusing on the choice of base pre-trained Transformer, we can see that RoBERTa outperforms BERT when making predictions for contextualized sentences. This holds true for both the Min Match (41.75 vs. 38.36) and Exact Match (23.54 vs. 20.02) configurations. We hypothesize that because RoBERTa is pre-trained on a larger corpus, it learns more robust word embeddings relative to BERT. In addition, RoBERTa removes the next sentence prediction loss, is trained for longer with a larger batch size and uses longer sequences of input data. The RoBERTa architecture is therefore able to predict the sentence level tags with a greater F1 score and also exhibits a good level of label coherence when predicting sequences of tags for consecutive sentences.

\autoref{table:final_results} also illustrates that aggregating IOBES predictions into initiatives leads to a performance improvement over binary predictions in the Exact Match context while sacrificing some performance in the Min Match setting regardless of the base Transformer model. For example, RoBERTa IOBES Without Context yields an Exact Match F1 of 18.72\% relative to 13.16\% for RoBERTa Binary Without Context while reducing F1 performance in the Min Match setting by 2.69\%. Manual inspection of the predictions reveals that using IOBES to define the precise initiative span yields results which are more closely aligned with the ground truth for multi-sentence initiatives. On the other hand, using binary predictions performs better when detecting single-sentence initiatives as there are more examples available to train the model compared to dividing up relevant sentences into IOBES classes. 

Based on the improved Exact Match performance we use the multi-class predictions in the final system, which also improves overall discourse coherence.
While this comes at the cost of missing some single-sentence initiatives, we prioritise performance on the fully autonomous system because it is typically more expensive to deploy a human-in-the-loop system that directs an annotator towards relevant sections of a report for manual inspection. If this is feasible however, the binary detection system may be more desirable; we leave this for future research. 

Finally, the results in \autoref{table:final_results} also illustrate that adding context helps to improve the F1 score measured at the initiative level across both sets of metrics. In particular, adding 1 sentence of symmetrical context on either side of the target sentence improves the F1 initiative level score on the test set by 4.8\% using the Min Match metric and 4.82\% using the Exact Match metric relative to the no-context RoBERTa IOBES model.

The final model hence uses pre-trained RoBERTa, takes multiple context sentences as input and predicts fine-grained multi-class IOBES labels. This system achieves the best performance and is able to predict the distinct span of each initiative more accurately. It outperforms a baseline binary BERT classifier by 7.98\% and achieves an F1 score of 23.54\% on the exact matching of initiatives. Using the partial matching metric, this system achieves results that are comparable to the BERT baseline (only 0.17\% lower). All the multi-class models improve recall while losing some precision, which is a result of the increased complexity in predicting the precise boundaries of an initiative compared to simply deciding whether a sentence is part of an initiative or not. The binary classifier thus maximizes its chance to detect single-sentence initiatives, however, it is less coherent when it comes to predicting multi-class initiatives. 


\subsection{Qualitative Evaluation}

In addition to a quantitative evaluation, we also carried out a qualitative evaluation and present some sample system output in \autoref{sec:appendix}. Based on this output, we see that the final model is able to successfully extract multi-sentence initiatives and produce coherent predictions for paragraphs of text referring to a given sustainability initiative. The main challenges are posed by Singleton initiatives, which are responsible for most of the false positive and false negative predictions. These examples emphasize the subjective nature of the task and support the idea that having an automated labeling process scales well across large bodies of text. After taking a closer look at the Singleton predictions that the model struggles to label correctly, it is also encouraging to note that many of the false positive Singleton predictions are plausible initiatives that may have been missed by the human experts due to annotation fatigue.


\section{Additional Training Data}

We also investigated how using additional training data affects the results. 
The public dataset we release is a subset of a larger private dataset of 1.5 million sentences from 507 company reports. In this section we report the effects of adding this additional data to the training process.
When the model is trained on this expanded training set, performance increases in terms of both the Min Match and Exact Match metrics. 
We also find that a larger dataset allows the model to better utilise a larger context window -- while having 1 sentence of context on either side of the target sentence was found to be optimal in the original training set, 2 sentences achieve a higher result using the extra data.
By having access to many more additional examples, the model is able to learn more complex patterns and relations that span across a larger number of sentences.

These observations show that there is room for model improvement using the public training set. Methods such as learning from noisy labels or optimization with additional multi-task objectives could help the model make better predictions without requiring millions of annotated sentences. Similarly, different architectures could be developed for more optimal cross-sentence contexts, potentially spanning a large number of sentences.

\begin{table}\centering
\begin{adjustbox}{width=0.48\textwidth}
\begin{tabular}{@{}lcc@{}}\toprule
RoBERTa Model & Min Match F1 & Exact Match F1\\
\midrule
IOBES 1 Sent Context & 46.84 & 26.96 \\
IOBES 2 Sent Context & \textbf{49.16} & \textbf{30.05} \\
\bottomrule
\end{tabular}
\end{adjustbox}
\caption{Initiative level F1 scores on the test set with varying levels of context. Models were trained on a larger, private training set of 507 company reports encompassing 1.5 million sentences.}
\label{table:add_training}
\end{table}


\section{Conclusion}

In this paper we have introduced the novel task of detecting sustainability initiatives in company reports. We release a new dataset of reports that have been manually annotated for mentions of sustainability initiatives. 
The task requires systems to make predictions at the sentence level and benefits from taking surrounding context sentences and labels into account.

We evaluate two pre-trained Transformer models on this task and also propose a novel architecture that specialises in contextual sentence classification. The model takes multiple sentences as input and encodes them using a Transformer layer. Multi-class IOBES labels are then predicted for each input sentence using a CRF -- this encourages the model to predict labels that have improved coherence with the surrounding labels while also providing a multi-task learning benefit. The predicted sentence-level labels are then combined into a list of initiatives which we evaluate against the gold-standard. The results show that both fine-grained IOBES labeling and additional context improve model performance on the task of predicting the exact span of each initiative.

Potential new areas of research include investigating the use of longer context windows to identify the sentence span of distinct initiatives, or event extraction methods which can perhaps be incorporated as complementary objectives in the main classification task. Finally, new research can explore using human-in-the-loop methods to fine-tune the annotation process and address the subjective nature of the task. Specifically, we found that binary sentence classification was an effective way of identifying the approximate location of initiatives in a report, which might be helpful in augmenting the human annotation process. 

\clearpage

\bibliography{acl_latex}
\bibliographystyle{acl_natbib}

\clearpage
\onecolumn

\appendix

\newcolumntype{C}[1]{>{\let\newline\\\arraybackslash\hspace{0pt}}m{#1}}

\section{Sustainability initiatives examples and guidelines}
\label{sec:guidelines}

As defined by the Leonardo Centre, an initiative is a statement of action connected to an identified sustainability objective. An initiative is not a mere declaration of intent or a statement of goal or objective, since it must imply a concrete
action. \autoref{table:initiative_example} illustrates a set of examples and counterexamples of potential sustainability initiatives included as part of the annotation guidelines. We provide an explanation for each record in the table in order to justify the annotation decision. 

The annotations were performed by PhD students at \citet{leonardo2014} as part of their research projects. All reports in the dataset are publicly available on each company's website and we provide a script to automatically download these and map them against the official labels. \\ 

\begin{table*}[h]\centering
\begin{adjustbox}{width=\textwidth}
\begin{tabular}{C{15cm}C{3cm}C{4cm}}\toprule
\multicolumn{1}{c}{Paragraph}& 
\multicolumn{1}{c}{Sustainability Initiative Label} & \multicolumn{1}{c}{Explanation}\\
\midrule
In November 2011, Orange won the “Entreprise numerique eco-engagee” prize in the framework of awards that promote green digital technology and reward creation, invention and digital technology that promotes sustainable development. 
& \xmark & Initiative description provided, but lacks concrete initiative action.\\
\midrule
In November 2011, Orange won the “Entreprise numerique eco-engagee” prize for its Shared Medical Imaging solution. The Shared Medical Imaging enables users to render radiology images virtual, while ensuring that the images, radiologist’s notes, minutes and records of tests carried out remain accessible. Shared Medical Imaging ensures that all data can shared in a secure way between healthcare professionals in different establishments. & \checkmark & Prize award represents the initiative description while exemplifying the use of the Shared Medical Imaging solution represents the initiative action.\\
\midrule
Energy efficiency is a key component of our environmental management system and environmental policies. This year, we achieved an extra electricity saving of EUR 1.9 million in total. & \xmark & This extract simply states a goal without clarifying the company action taken to achieve it.\\
\midrule
Energy efficiency is a key component of our environmental management system and environmental policies. This year, we achieved an extra electricity saving of EUR 1.9 million in total, thanks to the further
virtualization of our server park. & \checkmark & Both initiative description and initiative action are made explicit. The electricity saving was possible due to the virtualization of the servers.\\
\midrule
The Audit Committee, elected by the General Assembly, is responsible for the follow-up and handling of the denunciations made through this system and for making them known to external and internal entities whose involvement is deemed mandatory or justified. & \xmark & Description of general company duties do not constitute initiatives.\\
\midrule
We provide nutrition information on nearly all product labels, with the exception of certain returnable bottles, fountain beverages and waters (unsweetened, unflavored). For these beverage and packaging types, nutrition information is provided by alternate means through our Company and Coca-Cola system websites and consumer hotlines to guide consumers to additional information. & \checkmark & Activities which generate awareness and provides critical information to stakeholders are considered as initiatives.\\
\bottomrule
\end{tabular}
\end{adjustbox}
\caption{Examples of annotation guidelines for sustainability initiatives}\label{table:initiative_example}
\end{table*}

\clearpage

\section{Sample final-model predictions on example report}
\label{sec:appendix}

\begin{table*}[h]\centering
\begin{adjustbox}{width=\textwidth}
\begin{tabular}{C{15cm}cc}\toprule
\multicolumn{1}{c}{Predicted Initiative Text Span}& \multicolumn{1}{c}{Prediction Type} & \multicolumn{1}{c}{Classification}\\
\midrule
Ethical and social audits Suppliers of Carrefour products agree to comply with Carrefour’s supplier charter, drawn up in partnership with the International Federation of Human Rights (FIDH).
& Singleton & True Positive\\
\midrule
International food collection For the second year running, the Carrefour Foundation worked with Food Banks in 10 countries to conduct a food drive in over 2,300 Carrefour stores. With the help of customers, employees and volunteers, the equivalent of over 42 million meals were donated, compared with 9 million in 2013. & BE & True Positive\\
\midrule
Carrefour continues to develop solutions based on a circular economic model and local waste management, with the goal of recovering 100\% of waste produced in stores. In 2014, the waste recovery rate increased by 3.5 points (vs 2013) to reach nearly 65\%, and the percentage of recycled organic waste rose in conjunction with the biomethanisation project (12\% of waste recycled in 2014, vs 9\% in 2013). & BE & True Positive\\
\midrule
Carrefour constantly works to integrate a growing number of people with disabilities into the workplace. Its policy centres on three priorities: recruitment/integration, training and retention. At the end of 2014, Carrefour employed more than 11,200 people with disabilities, an increase of 21.3\% over 4 years. & Outside & False Negative\\
\midrule
The measures go hand in hand with its on-going solidarity policy: products withdrawn from sale for reasons unrelated to quality are distributed to associations such as Food Banks, of which Carrefour is a leading private partner.
& Outside & False Negative\\
\midrule
With the company’s support, Carrefour employees take part in socially-responsible initiatives to reduce exclusion at local, national and international levels. & Outside & False Negative \\
\midrule
Carrefour is also committed to the protection of ﬁsh stocks, promoting ﬁsh sourced from ecocertiﬁed ﬁshing and aquaculture, while halting the sale of deep sea ﬁsh. 
& Singleton & False Positive\\
\midrule
Rejuvenation of Carrefour Quality Lines under the name Jakość z Natury Carrefour (quality from nature) and expansion of the offering with new product lines: traditional Osełka butter and free-range chicken. All Jakość z Natury Carrefour products are sourced from small local producers who practice sustainable farming. 
& BE & False Positive\\
\midrule
Carrefour has also supported the efforts of the From Nord ﬁshery in France to have its sole product line evaluated by MSC. 
& Singleton & False Positive\\
\bottomrule
\end{tabular}
\end{adjustbox}
\caption{Illustrative extract of final model predictions on example Carrefour 2014 report: \url{https://www.carrefour.com/sites/default/files/2020-01/Carrefour-annual-report-2014-EN_0.pdf}}
\label{table:predicted_initiatives}
\end{table*}

\end{document}